\documentclass{article}
\usepackage{iclr2027_conference,times}

\usepackage{amsmath,amsfonts}
\usepackage{algorithmic}
\usepackage{algorithm}
\usepackage{array}
\usepackage{textcomp}
\usepackage{url}
\usepackage{verbatim}
\usepackage{graphicx}
\usepackage{booktabs}
\usepackage{enumitem}
\setlist[itemize]{leftmargin=*, topsep=0.5pt, partopsep=0.5pt,
                  itemsep=0.5pt, parsep=0.5pt}
\setlist[enumerate]{leftmargin=*, topsep=2pt plus 1pt, partopsep=0.5pt,
                    itemsep=1.5pt plus 1pt, parsep=1.5pt plus 1pt}
\usepackage{wrapfig}
\usepackage{xcolor}
\usepackage{tikz}
\usepackage{siunitx}
\usetikzlibrary{arrows.meta,positioning,calc,fit,backgrounds}
\usepackage[bookmarks=True]{hyperref}
\hypersetup{
    colorlinks=true,
    linkcolor=blue,
    citecolor=blue, 
    filecolor=magenta,
    urlcolor=cyan
    }
\newcommand{\sysname}{PanTact}

\makeatletter
\def\section{\@startsection{section}{1}{\z@}%
  {-4.0pt plus -0.5pt minus -0.5pt}
  {1.7pt plus 0.5pt minus 0.3pt}
  {\large\sc\raggedright}}
\def\subsection{\@startsection{subsection}{2}{\z@}%
  {-3.1pt plus -0.5pt minus -0.5pt}
  {0.5pt}
  {\normalsize\sc\raggedright}}
\makeatother

\makeatletter
\def\maketitle{\par
\begingroup
   \def\thefootnote{\fnsymbol{footnote}}
   \def\@makefnmark{\hbox to 0pt{$^{\@thefnmark}$\hss}}
   \long\def\@makefntext##1{\parindent\z@\noindent$\m@th^{\@thefnmark}$\,##1}
   \@maketitle \@thanks
\endgroup
\setcounter{footnote}{0}
\let\maketitle\relax \let\@maketitle\relax
\gdef\@thanks{}\gdef\@author{}\gdef\@title{}\let\thanks\relax}
\makeatother

\iclrfinalcopy

\title{\sysname{}: Teaching Whole-Body Dexterity
with a Wearable Contact Interface}

\author{
Xiaomeng Xu$^{1,2}$ \quad Han Zhang$^1$ \quad Zhian Ruan$^1$ \quad Xiaofeng Guo$^2$ \quad Jinkun Cao$^2$ \\
\textbf{Ruihan Yang$^2$ \quad
Carmelo Sferrazza$^2$ \quad Shuran Song$^{1,\dagger}$ \quad Guanya Shi$^{2,3,\dagger}$} \\[0.5em]
$^1$Stanford University \quad $^2$Amazon FAR \quad
$^3$Carnegie Mellon University \quad $^\dagger$Equal Advising
}

\begin{document}

\maketitle
\lhead{Preprint} 

\begin{abstract}
Humanoid robots have many degrees of freedom and contact surfaces, yet most loco-manipulation research restricts manipulation to the hands and avoids other body contacts with the environment. 
To extend manipulation beyond the hands, we present \sysname{}\footnote{The name \sysname{} combines \emph{pan-} (all) with \emph{tact-} (contact, tactile), reflecting a humanoid that manipulates with any body surface.}, a system that learns humanoid \emph{whole-body dexterity}, the capability to manipulate with any suitable body surface, directly from robot-free human demonstrations. Our key idea is to use contact as manipulation-centric information that transfers across embodiments. Contact locations and timing specify the interactions needed to accomplish a task, while the robot chooses how to coordinate its body to realize them within its own physical constraints. Specifically, \sysname{} captures, learns, and executes whole-body contact via: 1) a wearable, robot-free data collection interface that augments egocentric vision and bimanual UMI grippers with whole-body modular contact patches to record body contact locations and states; 2) a visuomotor policy that mitigates the egocentric observation gap and predicts whole-body contact goals; and 3) contact-aware whole-body control formulations that realize active contact goals using suitable robot surfaces. We deploy \sysname{} on a Unitree G1 humanoid without additional body contact sensors, demonstrating whole-body dexterity involving body support, pushing, sweeping, and load-sharing. 
Website: \href{https://pantact.github.io}{https://pantact.github.io}.
\end{abstract}


\begin{figure}[h]
    \centering
   \includegraphics[width=\linewidth]{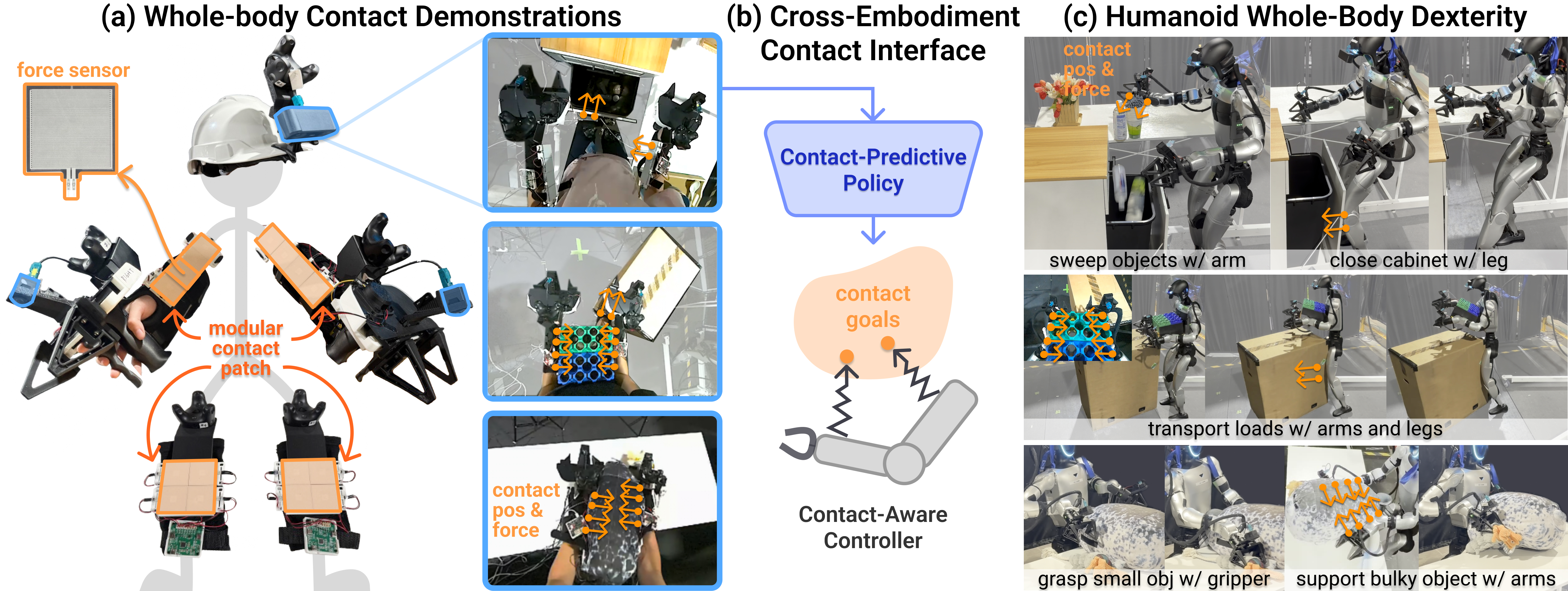}
   \vspace{-4mm}
    \caption{\textbf{\sysname{} gives a humanoid whole-body dexterity, learned from robot-free human demonstrations.} (a) The data collection interface captures whole-body contacts alongside vision and motion. (b) Contact provides an interface for human-to-robot transfer. The visuomotor policy predicts contact goals, and the whole-body controller realizes them using suitable robot surfaces. 
    (c) \sysname{} enables diverse whole-body dexterous skills on a humanoid. 
    }
    \label{fig:teaser}
\end{figure}

%
%
%

\section{Introduction}
\label{sec:intro}
When humans interact with the physical world, we use more than just our hands: our arms and torso embrace bulky objects, our forearms sweep multiple items or brace against surfaces, and our knees push objects when our hands are full. 
We call this capability \textbf{whole-body dexterity}~\citep{XuX-RSS-25}: the ability to accomplish manipulation tasks using any suitable surface of the body. 
Humanoid robots share the human body's overall morphology, with dozens of actuated degrees of freedom and potential contact surfaces distributed across the body. Yet much of loco-manipulation research limits manipulation to the hands, reserving the rest of the body for locomotion and balance and typically avoiding other body contact with the environment. This leaves much of the humanoid body's potential for manipulation unused. 

Realizing this potential requires coordinating whole-body motion with contact sequences across body surfaces. 
Human demonstrations provide natural examples of this coordination. We therefore ask: 
\textit{how can a humanoid learn whole-body dexterity directly from human demonstrations?}


Advances in imitation learning and demonstration interfaces~\citep{wu2024gello, cheng2025open, chi2024universal, xu2026hommi} have made learning manipulation from humans increasingly scalable. 
However, these pipelines capture and imitate either end-effector (EEF) or joint trajectories,  
and do not explicitly carry body-contact information from demonstration through policy prediction to robot execution. Extending these pipelines to whole-body dexterity introduces three connected challenges:
\begin{itemize}
    \item \textbf{Capturing body contact in human demonstrations.} Existing interfaces record only EEF motions or joint angles and do not directly measure where and when body contact occurs.
    \item \textbf{Preserving contact in human-to-robot policy transfer.} 
    EEF targets leave body contacts unspecified, while retargeting whole-body joints can alter contact locations because humans and robots differ in shape and proportions.
    \item \textbf{Executing contact with whole-body control.} Controllers designed to track motion references or EEF targets lack an explicit objective for achieving contact goals on the body.
    They must incorporate contact goals while maintaining motion tracking and balance.
\end{itemize}

To address the above challenges, our key observation is that contact provides \emph{manipulation-centric} information
that can be separated from embodiment-specific motion. Contact location and timing describe the task-level physical interaction, whereas the body surface and joint configuration used to produce it depend on the robot embodiment.
Contact also identifies which parts of the demonstrated body motion need to be preserved during transfer, allowing the robot to learn task-critical contacts without necessarily matching free-space body motions.
We therefore use \textbf{contact as a cross-embodiment interface}: \emph{measuring} contacts on the human, \emph{predicting} their desired states and locations with the policy, and \emph{tracking} active contact goals with the controller. 
Tracking only active contact goals without assigning them to particular robot links allows the controller to choose a body configuration suited to the robot to make contacts and maintain balance.
%

We present \sysname{}, a system that learns whole-body dexterity from robot-free human demonstrations and deploys it on an off-the-shelf humanoid through three components:
\begin{itemize}
    \item \textbf{Contact-complete, robot-free data collection system.} We augment egocentric sensing and bimanual UMI grippers with \emph{modular contact patches}, reconfigurable force-sensing tiles covering the demonstrator's forearms and knees that provide contact timing, intensity and location.
    \item \textbf{Contact-predictive whole-body manipulation policy.}
   The policy observes embodiment agnostic observations with body portions geometrically masked out from the egocentric view, and jointly predicts EEF trajectories and per-sensor contact goals, each specifying a contact state and position.
    \item \textbf{Contact-aware whole-body controller.} We propose three formulations to achieve active contact goals using suitable robot surfaces under physical constraints: upper-body contact-aware IK, whole-body IK executed by a dense tracker, and direct sparse contact tracking. 
\end{itemize}

We deploy \sysname{} on a Unitree G1 humanoid without adding body contact sensors. Our tasks combine grasping and body support to move differently sized objects simultaneously, sweeping objects with arm surfaces, using the legs for non-prehensile manipulation and locomotion at the same time, and sequentially switching contact across different body parts. Result videos are best viewed on \href{https://pantact.github.io}{https://pantact.github.io}.


\section{Related Work}
\label{sec:related}

\textbf{Whole-body sensing for robots.} Distributed range, force, tactile, and vision sensing has been used for collision avoidance, contact detection, and compliant whole-body interactions. Examples include range sensing for safe navigation and interaction~\citep{tanaka2015range, qi2022safe}, force and pressure sensing for compliant contact~\citep{kollmitz2018whole, goncalves2022punyo, xu2025compliant}, tactile sensing for contact feedback~\citep{liu2023enhancing, dean2019whole, jiang2024hierarchical, murooka2025tact, jang2026wtumi}, and whole-body vision on serial and soft growing robots~\citep{XuX-RSS-25, qin2026panovine}. 
These systems primarily integrate whole-body sensing into the robot and acquire demonstrations through teleoperation. \sysname{} incorporates whole-body tactile sensing into a wearable interface, and uses human contact measurements to supervise policy contact goal actions, which the robot realizes without additional body sensors.

\textbf{Data collection interfaces for robot learning.} 
Teleoperation provides robot-native demonstrations, but it is slow, costly, and hard to deploy on humanoids in diverse environments~\citep{wu2024gello, pmlr-v305-xiong25a}. UMI~\citep{chi2024universal} enables robot-free data collection with a portable handheld gripper, and its extensions add a head-mounted camera~\citep{xu2026hommi}, gripper-centric tactile and force sensing~\citep{choi2026inthewildcompliantmanipulationumift}, while humanoid-oriented interfaces additionally capture body or keypoint trajectories~\citep{nai2026humanoid}. 
These motion-centric interfaces capture EEF and keypoint trajectories or joint angles, but none of them measure body contact signals. 
\sysname{} adds contact sensing to preserve contact interactions during human to robot transfer.

\textbf{Learning loco-manipulation from demonstrations.}
A common architecture combines a high-level policy that predicts task-space commands with a whole-body controller that executes them~\citep{ nai2026humanoid, zhao2026halomi}. Controllers use optimization~\citep{ferrolho2023roloma, xu2026hommi} or learned motion tracking~\citep{he2024learning,  dugar2024learning, weng2025hdmi}, with demonstrations obtained from wearable devices~\citep{nai2026humanoid, zhao2026halomi}, human video~\citep{bahety2025safemimic}, or simulation~\citep{li2025momagen}, but the policy-to-controller interface typically specifies only motion targets. 
Recent work introduces contact-conditioned motion tracking~\citep{chen2026scenebot, li2026contactmimic}, but assigns contacts to predefined body parts and does not integrate contact prediction by a high-level visuomotor policy.
\sysname{} integrates human contact sensing, a visuomotor policy that predicts contact goals, and contact-aware whole-body control into a complete system.

\section{\sysname{} Data Collection Interface}
\label{sec:interface}

\begin{wrapfigure}{r}{0.6\linewidth}
    \vspace{-\intextsep}
    \centering
    \includegraphics[width=\linewidth]{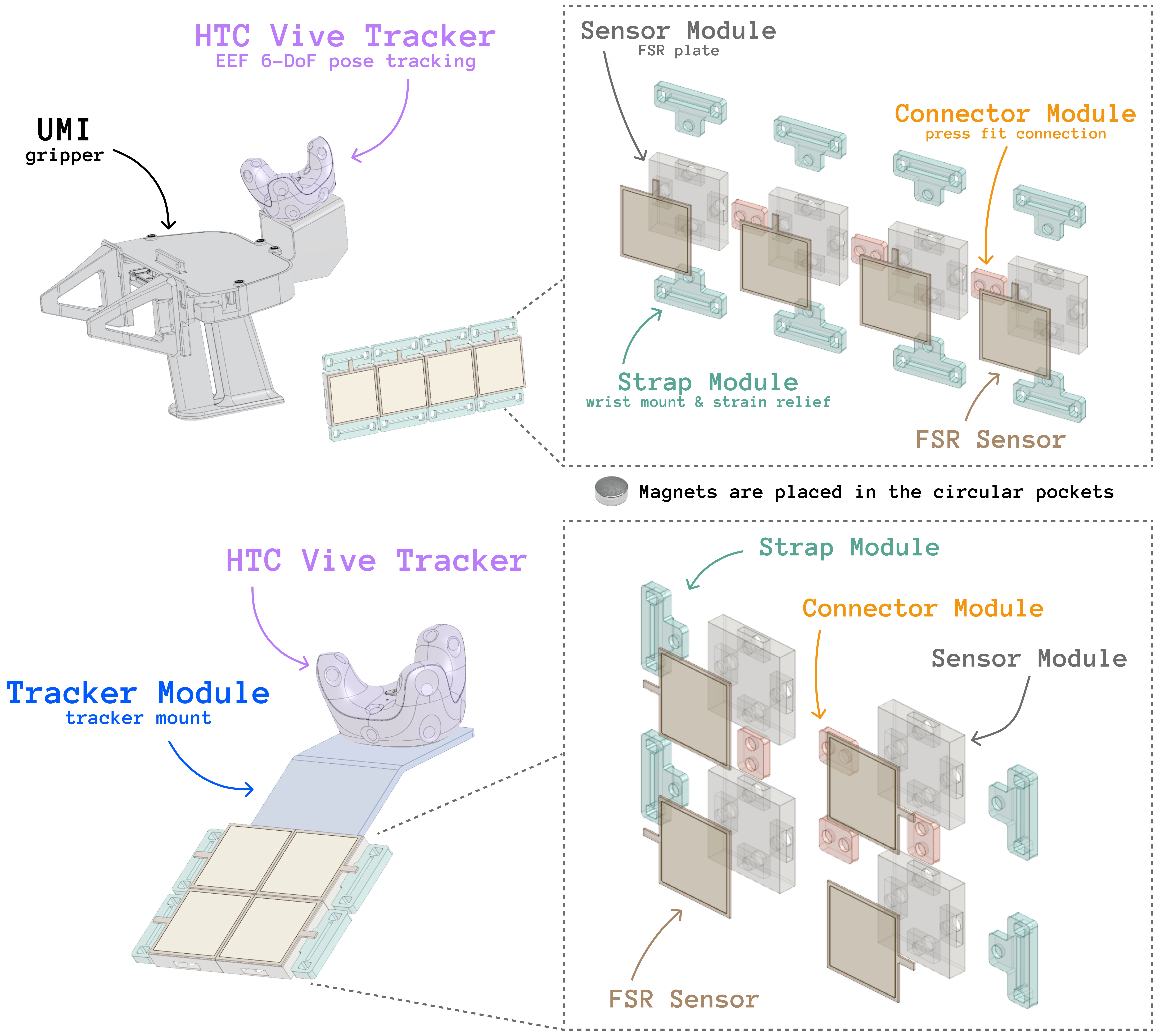}
    \vspace{-5mm}
    \caption{\textbf{\sysname{} modular contact patch.} Sensor, connector, and strap modules snap together into patches of arbitrary shape and extent. We showcase arm and knee patches.
    }
    \vspace{-4mm}
    \label{fig:hardware}
\end{wrapfigure}

\sysname{}'s data collection interface combines egocentric-augmented bimanual UMIs~\citep{chi2024universal,xu2026hommi} with modular contact patches (Fig.~\ref{fig:hardware}), capturing vision and motion together with body-contact signals that identify where and when contact occurs.
%
We record images from a head-mounted stereo camera (\texttt{ZED X Mini}) and two fisheye wrist cameras (\texttt{ZED X One S}) at \SI{60}{fps}, poses from HTC Vive trackers on the head and both wrists at \SI{250}{Hz}, and gripper widths at \SI{1}{kHz}. All streams share a monotonic clock and are aligned to \SI{60}{Hz} during post-processing. 

Existing whole-body tactile skins are typically tailored to a particular robot embodiment. To accommodate different demonstrators and body regions, we design the sensing surface from small, identical \textit{modules} that can be assembled into wearable \textit{patches} of varying shapes and sizes.

\textbf{Contact sensing and electronics.}
Each module is a $44\times44\,\mathrm{mm}$ unit carrying a single force-sensitive resistor (FSR).
A custom $50\times50\,\mathrm{mm}$ PCB simultaneously streams up to 16 FSR channels at $1\,\mathrm{kHz}$. Each FSR is connected in series with a fixed $1\,\mathrm{k}\Omega$ resistor to form a voltage divider, sampled by a 12-bit ADC with $8\times$ oversampling.
Modules communicate over a daisy-chained, multi-drop RS-485 bus using a request-and-response protocol with CRC checks, and also support direct USB readout. 

\textbf{Modular magnetic tiling.} Each sensing module has a connector slot on each side, which accepts press-fit magnetic connectors, allowing modules to snap into patches of varying shapes and sizes. We design three connector types: a full connector that bridges adjacent sensing modules; a strap connector that secures the patch to the body and provides cable strain relief; and a tracker connector that mounts a HTC Vive tracker to track the patch's 6-DoF pose.

\textbf{Contact detection and localization.} FSR resistance drifts with temperature and mounting pressure, so we binarize contact relative to a per-channel adaptive baseline rather than an absolute threshold. 
%
Each patch has an associated tracker, and a rigid transform maps each sensor center into the tracking world frame.
At each synchronized timestep, we store an $N\times4$ contact array, where $N$ is the number of FSR sensors, and each row contains a sensor's binary contact state and 3D position. 
These sequences supervise the policy's contact-goal predictions (\S~\ref{sec:policy}), with the binary state indicating when the controller should track contact location (\S~\ref{sec:controller}).

\begin{figure}
    \centering
    \includegraphics[width=\linewidth]{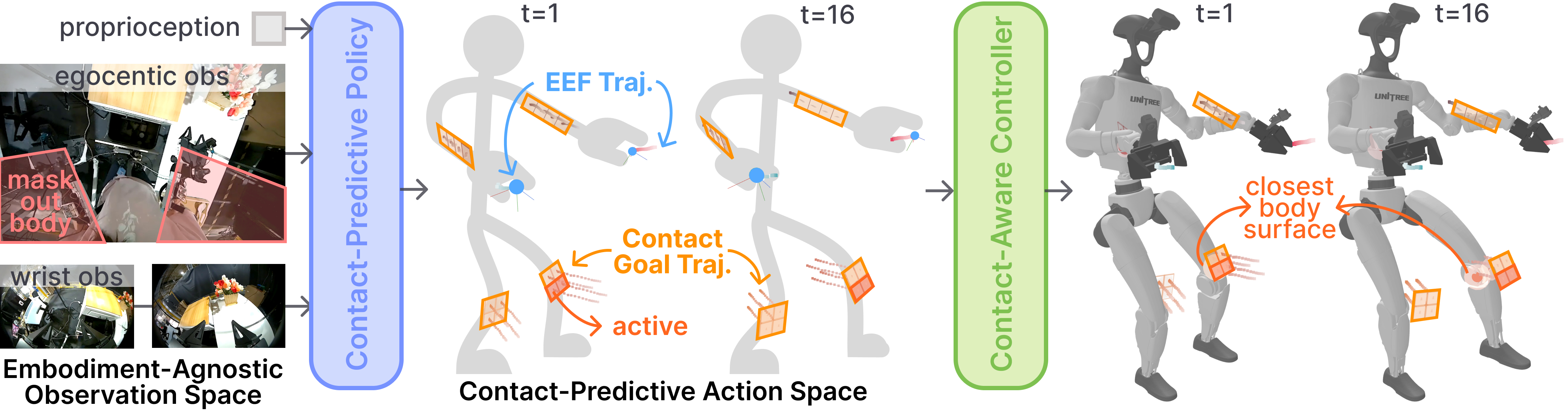}
    \vspace{-5mm}
    \caption{\textbf{Cross-Embodiment Contact Interface.} The \textit{Contact-Predictive Whole-Body Manipulation Policy} uses body-masked 3D egocentric observations, wrist images, and proprioception to predict EEF trajectories and contact goals (position and binary state). The \textit{Contact-Aware Whole-Body Controller} coordinates whole-body motion to track EEF commands and realize active contact goals using suitable robot surfaces.}
    \label{fig:policy}
\end{figure}

\section{Contact-Predictive Whole-Body Manipulation Policy}
\label{sec:policy}
To learn whole-body dexterity from human demonstrations, we employ a policy that predicts whole-body contact goals alongside EEF trajectories (Fig.~\ref{fig:policy}).
To mitigate the human-to-robot embodiment gap, the policy masks body regions from the egocentric view (\S~\ref{sec:policy:obs}) and predicts world-frame EEF and contact goals as actions (\S~\ref{sec:policy:action}).

\subsection{Embodiment-agnostic Observation space}
\label{sec:policy:obs}

Whole-body contacts often occur behind the grippers, outside the views of wrist cameras. An egocentric view provides broader scene context and observes whole-body interactions, but introduces a larger human-to-robot embodiment gap. During demonstrations, the demonstrator's body often appears in the egocentric view, creating a mismatch in appearance between training and deployment. Differences in human and robot height also produce a viewpoint mismatch.


Following HoMMI~\citep{xu2026hommi}, we use a 3D egocentric representation to reduce sensitivity to viewpoint differences and geometrically mask body regions. Each gripper frame and contact patch defines a body-facing half-space, since arms originate behind the grippers, and body surfaces start behind the contact patch surfaces.
We approximate the body region by the intersection of these half-spaces and mask the corresponding points in the attention layer of the egocentric vision encoder (Fig.~\ref{fig:policy}).
This construction requires no appearance model of either embodiment and scales to any number of contact patches.
The policy's overall observation space is
\begin{equation}
    \mathbf{o}_t = \left( P^{\mathrm{ego}}_{t-h+1:t},\; I^{L}_{t-h+1:t},\; I^{R}_{t-h+1:t},\; \mathbf{x}^{L}_{t-h+1:t},\; \mathbf{x}^{R}_{t-h+1:t} \right),
    \label{eq:obs}
\end{equation}
where $h{=}2$ is the number of history steps, $P^{\mathrm{ego}}$ is the masked egocentric pointmap, $I^{L},I^{R}$ are the wrist images, and $\mathbf{x}^{L}, \mathbf{x}^{R}$ each contain an EEF pose and gripper width. 

\subsection{Contact-Predictive Action Space}
\label{sec:policy:action}

We use a diffusion policy architecture~\citep{chi2025diffusion} that runs at \SI{10}{Hz} and predicts action chunks with a horizon of $H{=}16$ steps. Each step is
\begin{equation}
    \mathbf{a}_t = \Big( \boldsymbol{\lambda}_t,\; \mathbf{x}^{L}_t,\; \mathbf{x}^{R}_t,\; \{ (\hat{c}_{t,i},\; \hat{\mathbf{p}}_{t,i}) \}_{i=1}^{N} \Big),
    \label{eq:action}
\end{equation}
comprising a 3D look-at point $\boldsymbol{\lambda}_t$ for the head~\citep{xu2026hommi}, target EEF poses and gripper widths $\mathbf{x}^{L}_t,\; \mathbf{x}^{R}_t$, and \emph{contact goals} $\{ (\hat{c}_{t,i},\; \hat{\mathbf{p}}_{t,i}) \}_{i=1}^{N}$.
$\hat{c}_{t,i}$ is a binary contact state that determines whether the contact position $\hat{\mathbf{p}}_{t,i}$ is a tracking target: only active goals ($\hat{c}_{t,i}=1$) request the robot to make contact with the world at $\hat{\mathbf{p}}_{t,i}$. Inactive goals ($\hat{c}_{t,i}=0$) do not require the robot to reproduce the corresponding sensor's free-space trajectory. 
The whole-body controller (\S~\ref{sec:controller}) selects a suitable configuration to achieve the intended active contacts while not being over-constrained by inactive contacts.
We express the pointmaps, proprioception poses, EEF poses and contact positions across each observation history and action chunk in the same frame, relative to the latest observed left-gripper pose~\citep{xu2026hommi}, reducing sensitivity to differences in human and robot kinematics.


\section{Contact-Aware Whole-Body Control}
\label{sec:controller}

\subsection{General formulation}
\label{sec:controller:overview}

The whole-body controller converts the policy's EEF trajectories and contact goals $\{(\hat{c}_i,\hat{\mathbf{p}}_i)\}$ into robot whole-body motions. It tracks contact locations only when $\hat{c}_i=1$, prioritizing task-critical interactions without requiring the robot to reproduce demonstrator's free-space body motion. Each active goal can be realized by any suitable robot surface, leaving the controller to coordinate the remaining degrees of freedom under robot-specific constraints.

\textbf{Any-surface contact objective.}
Let $q$ denote the robot configuration, $\mathcal{B}$ the set of candidate robot bodies, and $\mathrm{d}_b(\mathbf{p},q)$ the distance from point $\mathbf{p}$ to body $b$'s collision surface. The proximate distance from a commanded contact $\mathbf{p}$ to the robot is:
\begin{equation}
    \mathrm{d}(\mathbf{p},q)
    = -T \log \sum_{b \in \mathcal{B}}
    \exp\!\big(-\mathrm{d}_b(\mathbf{p},q)/T\big),
    \label{eq:softmin}
\end{equation}
where $T=0.02\,\mathrm{m}$ controls the smoothing. 
The whole-body controller drives $\mathrm{d}(\hat{\mathbf{p}}_i, q) \to 0$ for active contacts to approach the target.

\textbf{Virtual contact targets.}
Reaching a contact location alone does not prescribe a pressing force.
Following the virtual-target principle~\citep{hou2025adaptive}, we use a \emph{virtual contact target}: a controller setpoint displaced beyond the intended contact location into the object. When the object constrains the robot surface at contact, the residual tracking error generates a pressing effort through the low-level controller.
At deployment, we offset active contact targets by:
\begin{equation}
    \tilde{\mathbf{p}}_i
    = \hat{\mathbf{p}}_i + \hat{c}_i\,\delta\,\mathbf{n}_i,
    \qquad
    \delta = \min\!\left(
        \frac{F_{\mathrm{des}}}{k_{\mathrm{virt}}},
        \delta_{\max}
    \right),
    \label{eq:virtual}
\end{equation}
where $\mathbf{n}_i$ is a unit pressing direction into the object, $F_{\mathrm{des}}$ is a nominal per-site pressing force, $k_{\mathrm{virt}}$ is a design stiffness, and $\delta_{\max}$ bounds the offset. The controller tracks $\tilde{\mathbf{p}}_i$ to encourage pressing force.

\subsection{Whole-Body Control Strategies}
\label{sec:controller:formulation}
We implement contact-aware tracking objective through three formulations: an upper-body IK, whole-body IK executed by an existing dense motion tracker, and a sparse contact tracker that we train to follow EEF and contact goals directly.

\noindent\textbf{I. Upper-Body IK.}
\label{sec:controller:ik}
A straightforward solution is to add contact tasks to differential IK. At each step, the solver jointly optimizes EEF and contact tracking through the quadratic program (QP):
\begin{equation}
    \min_{\dot{q}} \;\;
    \sum_{k \in \mathcal{T}} w_k
    \big\|J_k(q)\dot{q}-K_k e_k(q)\big\|^2
    + \varepsilon\|\dot{q}\|^2
    \quad \text{s.t.} \quad \dot{q}\in\mathcal{C},
    \label{eq:ik}
\end{equation}
where $q$ is the joint configuration and $\dot{q}$ is the commanded joint velocity. For task $k$, $e_k$ is the target-minus-current error, $J_k$ maps joint velocities to task-space velocities, $K_k$ is a feedback gain, and $w_k$ sets the task weight. The task set $\mathcal{T}$ includes EEF position and orientation tracking, nominal-posture regularization, and active contact tracking. The coefficient $\varepsilon>0$ penalizes joint velocity, and $\mathcal{C}$ enforces joint-position and velocity limits.
For each active contact $\hat{\mathbf{p}}_i$, the solver finds the closest surface point $\mathbf{s}_b$ on each candidate body and retains the three nearest bodies. Their contact-task weights are proportional to $\exp(-\mathrm{d}_b/T)$ and normalized over the selected bodies. Each receives a position task that brings $\mathbf{s}_b$ toward the target. 
%
The resulting joint velocities are integrated into joint-position targets for the arms, while a standing controller controls the legs. This formulation requires neither training data nor simulation, but is limited to upper-body tasks with fixed legs.

\noindent\textbf{II. WBIK + SONIC.}
\label{sec:controller:wbik}
To extend contact tracking to the lower body, we expand the QP to optimize whole-body degrees of freedom and use the existing SONIC motion tracker~\citep{luo2026sonic} for execution. When the EEF and contact goals are reachable within the current stance, the feet remain planted while the solver adjusts posture through squatting and leaning. Otherwise, it commands a stepping motion. The resulting full-body configuration trajectory provides a dense reference that SONIC executes by tracking all joint targets.
This formulation enables lower-body contacts and locomotion while reusing an existing whole-body tracking controller. However, the dense reference specifies postural details that are not essential to the task, and tracking them can compete with EEF and contact accuracy. Moreover, the IK solver does not model SONIC's execution limits and may generate dynamically infeasible references.

\noindent\textbf{III. Sparse Contact Tracker.}
\label{sec:controller:rl}
To preserve freedom in whole-body coordination during execution, we train a controller that directly tracks sparse EEF and contact goals without an intermediate dense reference. Following masked motion tracking~\citep{tessler2024maskedmimic}, the controller learns to satisfy the provided targets while choosing the remaining body motion.

\textit{Controller interface.}
The tracker runs at $50\,\mathrm{Hz}$ and outputs 29 joint-position targets for PD control. It receives proprioception including joint positions and velocities, root orientation and angular velocity, and the previous action, together with a goal block containing masked future joint targets, EEF targets, and $S$ contact slots. Each slot contains a 3D point in the robot root frame and a contact state flag, with a separate validity mask identifying the slots to use. At deployment, the controller receives EEF targets and active contact goals, leaving other targets masked.

\textit{Hindsight contact labeling.}
 Training such a controller requires (goal, motion) pairs with achievable contact goals, which standard motion-capture data do not directly provide. 
 We construct these pairs by rolling out reference motions with an expert tracker in simulation and sampling target points uniformly over contactable body collision surfaces along the achieved trajectories.
 We apply this labeling to both a general motion-capture dataset used to train SONIC~\citep{luo2026sonic} for broad whole-body motion coverage, and to our human demonstrations, executed with WBIK+SONIC (\S~\ref{sec:controller:wbik}) under trajectory perturbations for task-specific coverage.
 During training, each episode draws $S$ slots from the clip's pool with a random number of active slots, and each selected slot is randomly marked as in-contact (to be tracked tightly) or pre-contact (to be approached softly). Contact targets derived from achieved rollouts thus become the commands the policy learns to satisfy, providing supervision without human annotation and exposing the controller to diverse contact configurations.

\textit{Training objective.}
The reward combines motion tracking, contact tracking, and regularization:
$r=r_{\mathrm{track}}+r_{\mathrm{contact}}+r_{\mathrm{reg}}$.
For the set $\mathcal{V}$ of valid training slots, the contact reward is
\begin{equation}
    r_{\mathrm{contact}}
    = w \sum_{i\in\mathcal{V}}
    \left[
        0.3\,c_i e^{-\mathrm{d}_i^2/\sigma_a^2}
        + 0.1\,(1-c_i)e^{-\mathrm{d}_i^2/\sigma_s^2}
    \right],
    \label{eq:rcontact}
\end{equation}
where $\mathrm{d}_i=\mathrm{d}(\hat{\mathbf{p}}_i,q)$ and $c_i$ distinguishes in-contact from pre-contact training goals. We use $\sigma_a=0.05\,\mathrm{m}$ for tight tracking and $\sigma_s=0.25\,\mathrm{m}$ for soft approach. The weight $w$ increases during training to introduce contact tracking after basic motion-tracking competence. The term $r_{\mathrm{track}}$ rewards body position, orientation, and velocity tracking alongside EEF tracking, while $r_{\mathrm{reg}}$ penalizes abrupt actions, joint-limit violations, and uncommanded contact.

We train with PPO~\citep{schulman2017proximal} in 4096 parallel simulated environments using hindsight-labeled rollouts generated by SONIC~\citep{luo2026sonic}, with domain randomization over masses, friction, and contact-goal perturbations. This formulation learns whole-body contact realization directly, at the cost of simulation training and dependence on the training distribution.
\section{Evaluation}
\label{sec:eval}
We evaluate whether whole-body dexterity can be learned from robot-free human demonstrations and transferred to a humanoid. Our evaluation probes four capabilities:
\begin{itemize}
    \item \textbf{Cross-embodiment contact transfer:} contacts measured on a human body and predicted by the visuomotor policy must be realized by a humanoid with a different configuration. This capability is required for all tasks.
    \item \textbf{Non-prehensile upper-body manipulation:} supporting bulky objects with the arms (\S~\ref{sec:eval:ulnar},~\ref{sec:eval:freight}) and sweeping objects with the arm surfaces (\S~\ref{sec:eval:tidying}).
    \item \textbf{Lower-body loco-manipulation:} coordinating the lower body for non-prehensile manipulation, locomotion, and balance at the same time (\S~\ref{sec:eval:tidying},~\ref{sec:eval:freight}).
    \item \textbf{Whole-body contact sequencing:} establishing, maintaining, and releasing contacts across different body regions as each task progresses, required for all tasks.
\end{itemize}
We compare against two pose-based interfaces that lack explicit body-contact objectives. These baselines test whether EEF trajectories alone, or with additional lower-body pose targets, suffice to transfer the demonstrated interactions without contact information:
\begin{itemize}
    \item \textbf{Humanoid UMI}~\citep{xu2026hommi,zhao2026halomi}: vanilla humanoid UMI setup, where a visuomotor policy predicts EEF actions from egocentric pointmap and wrist images and proprioception, and a whole-body controller tracks only EEF trajectories.
    \item \textbf{Humanoid UMI++}: an extension that additionally predicts and tracks knee poses. Each pose target is tied to a fixed robot body part and is tracked continuously.
\end{itemize}
All hardware experiments use a 29-DoF Unitree G1 equipped with grippers and cameras mirroring the data collection interface. For each task, all methods are trained on the same demonstrations and evaluated from matched test configurations.

\begin{figure}[t]
    \vspace{-\intextsep}
    \centering
    \includegraphics[width=\linewidth]{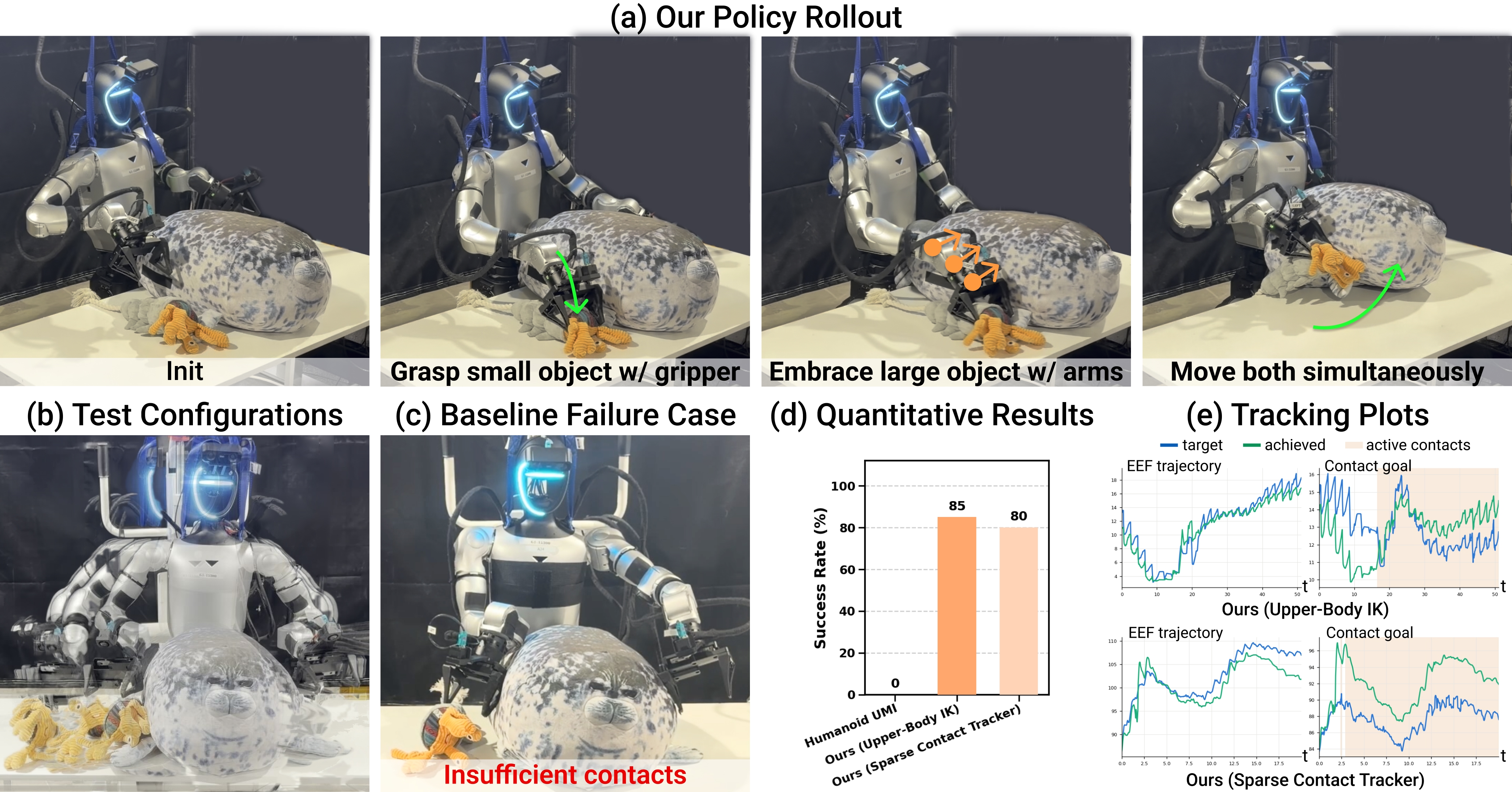}
    \caption{\textbf{Grasp-and-Embrace.}
    (a) Autonomous rollout of our policy deployed with \texttt{Upper-Body IK}, demonstrating capability to support a bulky object with arms while grasping small object with a gripper. (b) Overlayed test configurations with varied object placements and initial robot configurations. (c) Typical failure case of \texttt{Humanoid UMI} is insufficient arm contact to support the bulky object. (d) Task success rates. (e) Real-world tracking plots.
    }
    \label{fig:ulnar}
\end{figure}

\subsection{Grasp-and-Embrace}
\label{sec:eval:ulnar}
\textbf{Task:} 
The robot first grasps a small object from a table, then embraces a bulky object with its arms (Fig.~\ref{fig:ulnar}a). Success requires moving both objects simultaneously. 
\textbf{Capability:} Grasping the small object requires accurate EEF tracking, while supporting and moving the bulky object requires sufficient arm contact force on both sides.
\textbf{Data Collection:} We collected 100 demonstrations with varied table and goal placements and initial body postures. 
\textbf{Test Scenarios:} We evaluate each method over 20 rollouts with varied initial object placements and robot postures (Fig.~\ref{fig:ulnar}b).
%
\textbf{Performance:} \texttt{Ours (Upper-Body IK)} achieves an 85\% success rate, and \texttt{Ours (Sparse Contact Tracker)} achieves 80\% (Fig.~\ref{fig:ulnar}d). Occasional failures occur when the gripper misses the small object. Fig.~\ref{fig:ulnar}e shows accurate EEF and active-contact tracking with \texttt{Upper-Body IK}, while \texttt{Sparse Contact Tracker} exhibits slightly larger errors and tracking dead zones.
\texttt{Humanoid UMI} baseline consistently fails to support the bulky object because its whole-body controller tracks only EEF commands, leaving arm posture unspecified (Fig.~\ref{fig:ulnar}c). We also observe the grippers continuing reaching forward until approaching arm joint limits, potentially reflecting accumulated tracking error that takes the observation outside policy's training distribution.

\begin{figure}[t]
    \vspace{-\intextsep}
    \centering
    \includegraphics[width=\linewidth]{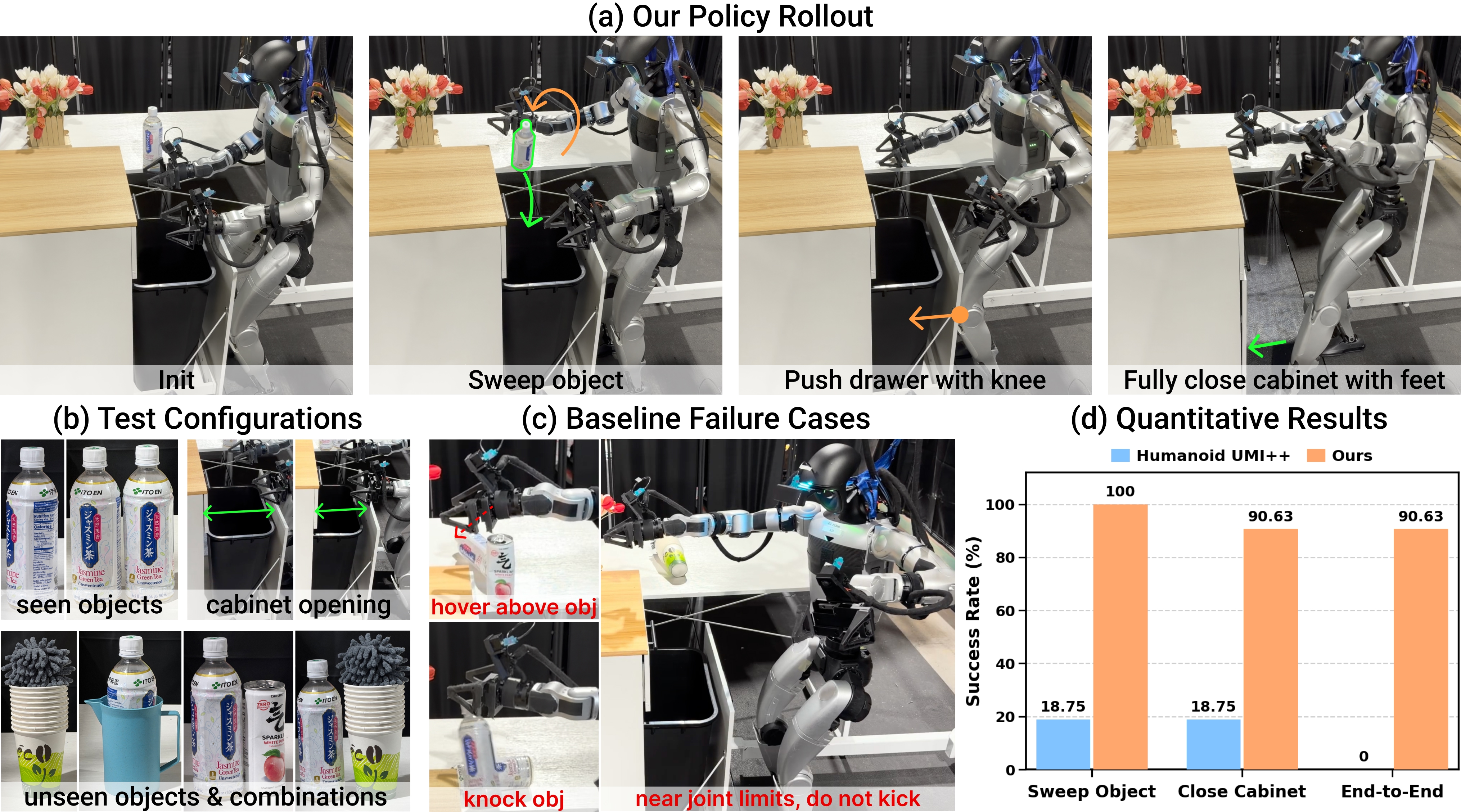}
    \caption{\textbf{Sweep-and-Stow.}
    (a) Our policy rollout with \texttt{WBIK+SONIC} tracker, demonstrating arm sweeping and lower-body pushing and locomotion at the same time.
    (b) Test configurations with seen and unseen object combinations, varied cabinet openings, and different robot-to-cabinet distances.
    (c) Typical failure case of the \texttt{Humanoid UMI++} baseline include the gripper passes above an object or knocks it outside the bin, and the robot fails to advance a leg for cabinet closure as its arms approach joint limits.
    (d) Stage-wise and end-to-end success rates.
    }
    \label{fig:stow}
\end{figure}
\subsection{Sweep-and-Stow}
\label{sec:eval:tidying}

\textbf{Task:} The robot uses arm to sweep one or more tall objects from a counter into an open bin, then uses a lower-body surface to push the cabinet closed (Fig.~\ref{fig:stow}a). The \textit{Sweep Object} score is based on the number of objects deposited in the bin. The \textit{Close Cabinet} score awards 50\% credit for partial closure below halfway and 100\% for closure of more than halfway. The end-to-end score is the product of the two stage scores. 
\textbf{Capability:} Sweeping requires coordinated arm contacts to guide objects into the bin, while cabinet closure requires lower-body pushing coordinated with stepping and balance. Across stages, the policy must predict contact goals appropriate to the current task stage, and the controller must select suitable surfaces to realize them.
%
\textbf{Data Collection:} We collected 150 demonstrations with varied object counts, types, and layouts, initial body postures, and cabinet openings. 
\textbf{Test Scenarios:} We evaluate each method over 16 rollouts, divided equally among four conditions: a single seen object, two seen objects, a single unseen object, and two unseen objects (Fig.~\ref{fig:stow}b). The seen-object conditions vary object and cabinet layouts. The single-unseen-object condition includes two objects with two rollouts each, the two-unseen-object condition includes two object pairs with two rollouts each.
%
\textbf{Performance:} \texttt{Ours (WBIK+Dense Tracker)} achieves an end-to-end success rate of 90.63\%. 
The robot sweeps with its wrist, forearm, or gripper side and closes the cabinet with either knee or a foot, adapting beyond the demonstrated forearm and left-knee contacts.
Occasional partial closures occur when the controller prioritizes balance over reaching the contact target. 
\texttt{Humanoid UMI++} often misses objects or knocks them outside the bin due to the lack of arm regularization. The arms also approach joint limits more often, accompanied by failures to advance a leg and close the cabinet (Fig.~\ref{fig:stow}c). This may be due to differences in human and robot proportions, as jointly tracking demonstrated hand and knee poses can require relative placements that are difficult for the robot to realize.

\subsection{Carry-and-Push}
\label{sec:eval:freight}

\begin{figure}[t]
    \vspace{-\intextsep}
    \centering
    \includegraphics[width=\linewidth]{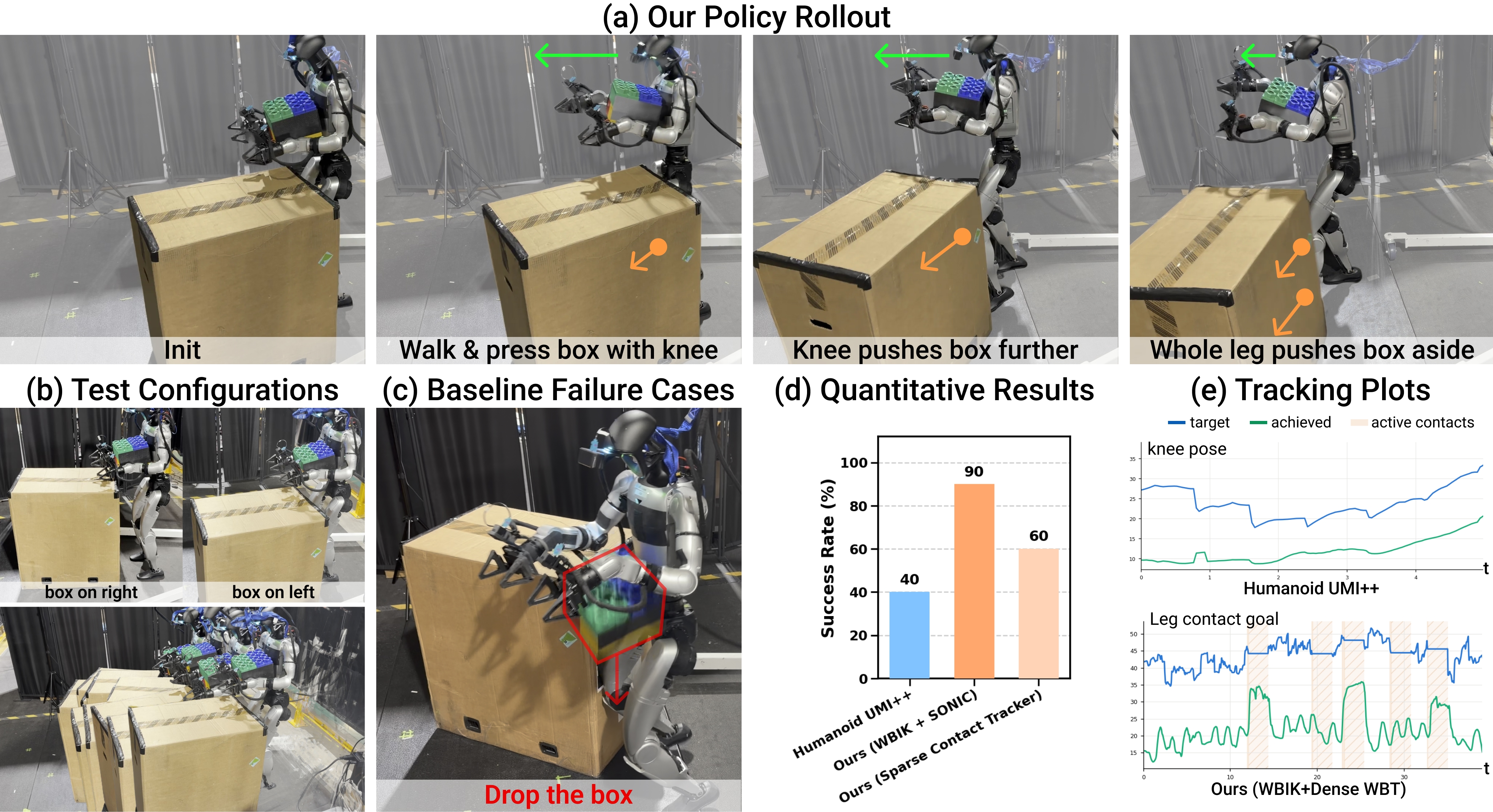}
    \caption{\textbf{Carry-and-Push.}
    (a) Our policy with \texttt{WBIK+SONIC} tracker, demonstrating arms load-sharing while walking and pushing with lower body.
    (b) Test scenarios with varied box placements and robot postures.
    (c) Failure case of \texttt{Humanoid UMI++}, dropping the carried box because of insufficient contact support. 
    (d) Task success rates.
    (e) Tracking plots of real-world rollouts.
    }
    \label{fig:box}
\end{figure}

\textbf{Task:} The robot carries a box in both arms, walks towards a larger box on the floor, and pushes it with its lower-body (Fig.~\ref{fig:box}a). Success requires retaining the carried box while moving the floor box forward by more than \SI{20}{cm}.
\textbf{Capability:} The arms must maintain sufficient contact support throughout locomotion, while the legs coordinate walking, pushing, and balance.
\textbf{Data Collection:} We collected 110 demonstrations with varied starting positions and box locations. 
\textbf{Test Scenarios:} We evaluate each method over 10 rollouts, with the floor box initially to the robot's right in five rollouts and to its left in the other five, and varied initial robot configurations (Fig.~\ref{fig:box}b).
\textbf{Performance:} \texttt{Ours (WBIK+Dense Tracker)} achieves a 90\% success rate. Failures occur when the robot drops the carried box, which happens more frequently with \texttt{Ours (Sparse Tracker)} and \texttt{Humanoid UMI++}. As shown in Fig.~\ref{fig:box}e, \texttt{Humanoid UMI++} continuously tracks knee poses, whereas \texttt{Ours} imposes leg contact objectives only during active contact phases. The baseline exhibits substantial knee-pose error, while \texttt{Ours} controller follows the commanded leg contact locations more accurately during active contact phases.

\begin{wrapfigure}{r}{0.58\linewidth}
    \centering
    \includegraphics[width=\linewidth]{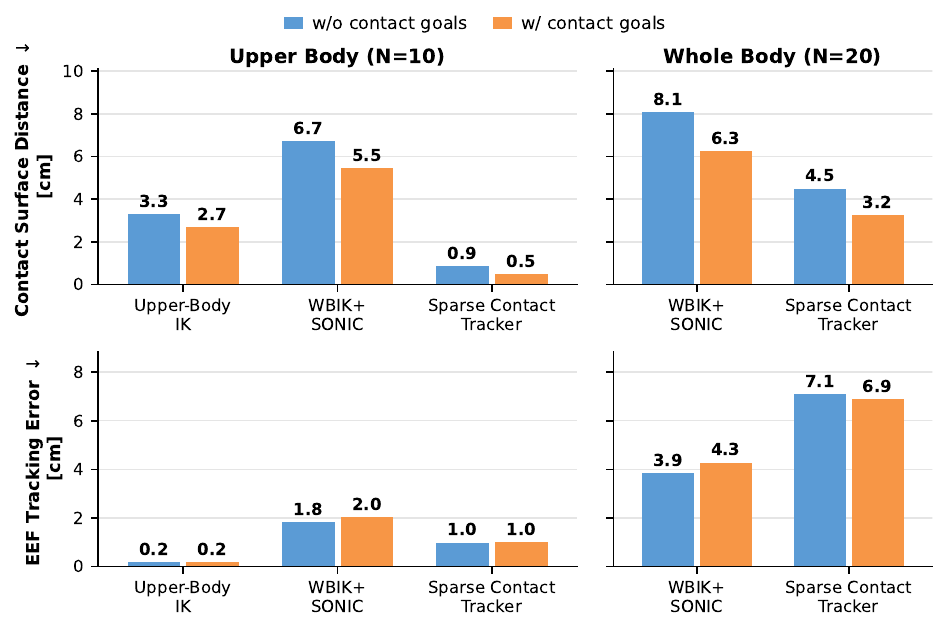}
    \vspace{-6mm}
    \caption{\textbf{Controller Comparisons.} We replay the demonstrations in simulation with our three whole-body control formulations, with and without contact goals.}
    \label{fig:results_controller}
\end{wrapfigure}

\subsection{Controller Comparisons}
\label{sec:eval:controller}
We replay the demonstration trajectories through the controllers in simulation, with and without contact goals. Fig.~\ref{fig:results_controller} reports contact surface distance, measured from each commanded contact point to the nearest robot surface, and EEF tracking error. 
We report medians of per-trajectory median errors over 10 Grasp-and-Embrace trajectories for upper-body evaluation, and 20 trajectories for whole-body evaluation, split equally between Sweep-and-Stow and Carry-and-Push.
Adding contact goals reduces contact surface distance for every controller, at little cost to EEF tracking accuracy. With contact goals enabled, \texttt{Upper-Body IK} achieves the lowest upper-body EEF error, while \texttt{Sparse Contact Tracker} achieves the smallest contact distance and outperforms \texttt{WBIK+SONIC} on both metrics. 
For whole-body tasks, the two trackers exhibit a tradeoff. \texttt{Sparse Contact Tracker} achieves a lower contact distance than \texttt{WBIK+SONIC}, but a higher EEF error.

\section{Conclusion}
\label{sec:conclusion}

We present \sysname{}, a system that learns humanoid whole-body dexterity from robot-free human demonstrations. The system treats contact as manipulation-centric information being transferred from human to robot with 1) modular contact patches for wearable, reconfigurable, low-cost body-contact sensing; 2) an embodiment-agnostic policy that predicts contact goals alongside end-effector trajectories; and 3) three whole-body controller formulations that realize active contact goals using suitable robot surfaces.
We hope the proposed hardware, learning framework, and controller formulations will encourage and facilitate future research on robot whole-body dexterity.

\section*{Acknowledgments}
The authors would like to thank Youjian Huang, Arnold Ong, Roberto Ceja, and Carrie Chen for their help with the G1 hardware setup. Ken Suzuki, Charlie Cheng and Umang Dighe for the UMI hardware setup.
Rocky Duan, Angjoo Kanazawa, Ruiqian Nai, Sihang Li, Yifan Hou, and Jinghan Sun for their helpful discussions.
This work was supported in part by Amazon, the NSF Award \#2143601, \#2037101, and \#2132519. The views and conclusions contained herein are those of the authors and should not be interpreted as necessarily representing the official policies, either expressed or implied, of the sponsors.

\bibliographystyle{iclr2027_conference} 
\bibliography{references}

\clearpage
\appendix
\section*{Appendix}

\section{Demonstration Interface Details}
\label{sec:app:interface}

\subsection{Recorded Data}

Each demonstration is a synchronized sequence
\begin{equation}
    \mathcal{D} = \Big\{ \big( I^{\mathrm{ego}}_t, P^{\mathrm{ego}}_t, I^{L}_t, I^{R}_t, T^{H}_t, \mathbf{x}^{L}_t, \mathbf{x}^{R}_t, \{ (c_{t,i}, f_{t,i}, \mathbf{p}_{t,i}) \}_{i=1}^{N} \big) \Big\}_{t=1}^{T_{\mathrm{demo}}},
    \label{eq:data}
\end{equation}
where $I \in \mathbb{R}^{H_{\mathrm{img}} \times W_{\mathrm{img}} \times 3}$ denotes RGB images and $P^{\mathrm{ego}} \in \mathbb{R}^{H_{\mathrm{img}} \times W_{\mathrm{img}} \times 3}$ contains a 3D point per egocentric pixel. The transform $T^{H}\in SE(3)$ denotes the head pose, and $\mathbf{x}^{L},\mathbf{x}^{R}$ each contains an EEF pose $T^{L},T^{R}\in SE(3)$ and gripper width $g^{L},g^{R}\in\mathbb{R}$. For sensor $i$, $c_{t,i}\in\{0,1\}$, $f_{t,i}\in\mathbb{R}_{\geq0}$, and $\mathbf{p}_{t,i}\in\mathbb{R}^{3}$ record binary contact, contact intensity, and location. Here, $N$ is the number of sensors and $T_{\mathrm{demo}}$ is the sequence length.

\subsection{Post-processing}
\textbf{Contact binarization.}
For sensor $i$, we compute the demonstrated contact state from its raw ADC count $r_{t,i}$ and running baseline $\beta_i$:
\begin{equation}
    c_{t,i} = \mathbf{1}\!\left[\, \beta_i-r_{t,i}
    > \max(0.35\,\beta_i,r_{\min}) \,\right],
    \label{eq:binarize}
\end{equation}
where $r_{\min}$ is a fixed noise floor.

\textbf{Visual processing.}
All cameras use a resolution of $960\times600$. We compute stereo depth offline using the neural depth model provided with the \texttt{ZED} camera, back-project it into a pointmap, and apply the body mask described in \S~\ref{sec:policy:obs}. 

\textbf{Synchronization.}
All sensor streams are logged on an \texttt{NVIDIA Jetson AGX Orin} and share a monotonic clock. During post-processing, we align them to the $60\,\mathrm{Hz}$ camera timeline, then sample synchronized sequences at \SI{20}{Hz} during training. Each training example uses the $h=2$ observation history and $H=16$ action horizon defined in \S~\ref{sec:policy}.

\section{Policy Details}
\label{sec:app:training}
We use a diffusion policy~\citep{chi2025diffusion} with a DiT denoising head (Fig.~\ref{fig:policy}), and finetune a pretrained DINO-v3 vision transformer~\citep{simeoni2025dinov3, xu2023jacobinerf} on egocentric and wrist images. For the masked egocentric pointmap, we augment image patch features with sinusoidal encodings of their 3D coordinates~\citep{xu2026hommi}. These observations condition the denoising transformer, trained with $\epsilon$-prediction and sampled with 16 DDIM~\citep{song2020denoising} steps at inference.
Tab.~\ref{tab:policy_hyperparameters} summarizes the training settings. We augment images with color jitter and use an exponential moving average (EMA) of the model weights for evaluation.

\begin{table}[h]
    \centering
    \footnotesize
    \caption{Policy architecture and training hyperparameters.}
    \label{tab:policy_hyperparameters}
    \begin{tabular}{@{}ll@{}}
        \toprule
        Parameter & Value \\
        \midrule
        Transformer depth / width / heads & 10 / 768 / 12 \\
        Training diffusion steps / inference DDIM steps & 100 / 16 \\
        Training epochs & 500 \\
        Effective batch size & 48 \\
        Base learning rate & $7.5\times10^{-5}$ \\
        Encoder learning rate & $7.5\times10^{-6}$ \\
        Learning-rate schedule & Cosine decay with warmup \\
        Validation split & $5\%$ \\
        \bottomrule
    \end{tabular}
\end{table}

\section{Controller Details}
\label{sec:app:controller}

We provide implementation details for the control strategies in \S~\ref{sec:controller}.

\subsection{IK and Reference Generation}
\label{sec:app:ik}

\textbf{Task weights and limits.}
Both upper-body and whole-body IK variants use Eq.~\eqref{eq:ik}, with EEF position and orientation weights of 500 and 200, a contact-task weight of 250, and a posture weight of 1 per joint group. Whole-body IK adds pelvis position weights $(80,80,40)$, orientation weights $(40,40,5)$, and a horizontal center-of-mass weight of 250, encouraging upright posture for stability while permitting greater variation in pelvis height and yaw. A velocity-damped configuration limit uses gain $0.95$, with joint-velocity bounds of $1\,\mathrm{rad/s}$ for the arms and waist and $2\,\mathrm{rad/s}$ for the legs.
At each iteration, we recompute the closest surface points, task weights, and Jacobians, then hold them fixed within the QP. 

\textbf{Stepping.}
In WBIK + SONIC, we impose hard SE(3) constraints on stance foot posture. The solver commands a step when the pelvis remains more than $0.12\,\mathrm{m}$ horizontally from the support midpoint for $0.2\,\mathrm{s}$. The solver releases the trailing foot's constraint and tracks a $0.4\,\mathrm{s}$ cycloidal swing with a peak lift of $0.06\,\mathrm{m}$. The stride follows the pelvis drift direction, with length clamped to $0.10$--$0.25\,\mathrm{m}$; swing position and orientation weights are 1000 and 100. During the swing, the stance foot remains fixed and the center-of-mass target shifts to it. At touchdown, the solver fixes the landed foot and enforces a $0.4\,\mathrm{s}$ cooldown before another step.

\subsection{Sparse Contact Tracker}
\label{sec:app:sparse}

\textbf{Contact slots.}
Upper-body trackers use $S=8$ contact slots, four per forearm; whole-body trackers use $S=16$, adding four per knee. The validity mask selects the slots included in $\mathcal{V}$ in Eq.~\eqref{eq:rcontact}. For each valid slot, $c_i=1$ requests tight in-contact tracking, while $c_i=0$ requests a softer pre-contact approach.
At episode reset during training, we sample the number of valid slots $|\mathcal{V}|$ uniformly from $\{0,\dots,S\}$ and draw their targets independently from the rollout's target pool. We permute the slot order and hold the sampled goals fixed throughout the episode.
At deployment, contact points are expressed in a gravity-aligned frame centered at the measured robot base and oriented by its yaw. 

\textbf{Reward.}
Tab.~\ref{tab:tracker_reward} lists the scales and weights for the reward defined in \S~\ref{sec:controller}. Motion-tracking terms use exponential kernels $e^{-e^2/\sigma^2}$, where $e$ denotes the corresponding global-frame error. EEF terms measure errors at the gripper tool-center points, with additional narrow kernels rewarding precise position and orientation tracking.

\begin{table}[t]
    \centering
    \footnotesize
    \caption{Sparse contact tracker reward terms.}
    \label{tab:tracker_reward}
    \begin{tabular}{@{}llcc@{}}
        \toprule
        Group & Term & $\sigma$ & Weight \\
        \midrule
        Pelvis & position / orientation & $0.3$ / $0.4$ & $1.0$ / $0.5$ \\
        Body & position / orientation & $0.3$ / $0.4$ & $1.0$ / $1.0$ \\
        Body & linear / angular velocity & $1.0$ / $3.14$ & $1.0$ / $1.0$ \\
        EEF & position / orientation & $0.3$ / $0.4$ & $1.2$ / $1.2$ \\
        EEF & linear / angular velocity & $1.0$ / $3.14$ & $1.2$ / $1.2$ \\
        EEF fine & position / orientation & $0.06$ / $0.1$ & $0.3$ / $0.3$ \\
        Contact & Eq.~\eqref{eq:rcontact} & -- & $1.0$ \\
        Regularization & action rate & -- & $-0.1$ \\
        Regularization & soft joint limit ($0.9$ of range) & -- & $-10$ \\
        Regularization & uncommanded contact & -- & $-0.1$ \\
        \bottomrule
    \end{tabular}
\end{table}

\textbf{Domain randomization.}
We randomize static friction over $[0.3,1.6]$, dynamic friction over $[0.3,1.2]$, restitution over $[0,0.5]$, and base center-of-mass offsets up to $(2.5,5,5)\,\mathrm{cm}$. Joint zero-position biases vary by $\pm0.015\,\mathrm{rad}$, PD gains by $\pm10\%$, and link masses by factors of $[0.8,1.5]$, with an additional base-mass offset of $[-1,3]\,\mathrm{kg}$. Pelvis velocity perturbations of up to $0.4\,\mathrm{m/s}$ occur every $1.5$--$3\,\mathrm{s}$.

\textbf{PPO settings.}
We train with PPO~\citep{schulman2017proximal} for 30,000 iterations in 4096 parallel environments. The actor learning rate starts at $2\times10^{-5}$ and adapts within $[10^{-5},2\times10^{-4}]$ using a KL target of $0.01$; the critic learning rate is $10^{-4}$. The entropy coefficient is $0.013$. The gradient norm clipping threshold is $0.1$, and the initial action noise is $0.05$.





\subsection{Real-world Deployment}
\label{sec:app:async}

Policy inference and low-level control run asynchronously. Each predicted action chunk is timestamped at its source observation. A scheduler discards expired steps, replaces overlapping targets with the latest prediction, and resamples commands to the tracker's $50\,\mathrm{Hz}$ control rate. 


The policy expresses actions in the left-gripper frame at the source observation. For execution, we transform them into the world frame using pose estimates from Super Odometry~\citep{zhao2021super}, which leverages the G1's built-in head-mounted LiDAR-inertial unit.

\end{document}